\documentclass[preprint]{vgtc}               % preprint
\ifpdf%                                % if we use pdflatex
  \pdfoutput=1\relax                   % create PDFs from pdfLaTeX
  \usepackage{graphicx}                % allow us to embed graphics files
  \DeclareGraphicsExtensions{.pdf,.png,.jpg,.jpeg} % for pdflatex we expect .pdf, .png, or .jpg files
\else%                                 % else we use pure latex
  \usepackage{graphicx}                % allow us to embed graphics files
  \DeclareGraphicsExtensions{.eps}     % for pure latex we expect eps files
\fi%

\graphicspath{{figures/}{pictures/}{images/}{./}} % where to search for the images

\usepackage{microtype}                 % use micro-typography (slightly more compact, better to read)
\PassOptionsToPackage{warn}{textcomp}  % to address font issues with \textrightarrow
\usepackage{textcomp}                  % use better special symbols
\usepackage{mathptmx}                  % use matching math font
\usepackage{times}                     % we use Times as the main font
\usepackage{cite}                      % needed to automatically sort the references
\usepackage{tabu}                      % only used for the table example
\usepackage{booktabs}                  % only used for the table example
\usepackage{enumitem}
\usepackage{graphicx}
\usepackage{bigstrut}
\usepackage{makecell}
\usepackage{booktabs} % 需要额外添加
\usepackage{array}
\usepackage{amsmath}
\usepackage{float}
\usepackage{pifont}
\usepackage{color,xcolor}
\usepackage[T1]{fontenc}

\vgtccategory{Research}

\vgtcinsertpkg

\preprinttext{Author’s preprint version. To appear in the proceedings of the IEEE VR 2024 Conference. It will be updated with a DOI when available.
}

\title{CMC: Few-shot Novel View Synthesis via Cross-view Multiplane Consistency}

\author{Hanxin Zhu\thanks{e-mail: hanxinzhu@mail.ustc.edu.cn}\\ %
        \scriptsize University of Science and Technology of China %
\and Tianyu He\thanks{e-mail: tianyuhe@microsoft.com}\\ %
     \scriptsize Microsoft Research Asia
\and Zhibo Chen\thanks{e-mail: chenzhibo@ustc.edu.cn}\\ %
     \scriptsize University of Science and Technology of China}

\teaser{
  \centering
  \includegraphics[width=\linewidth]{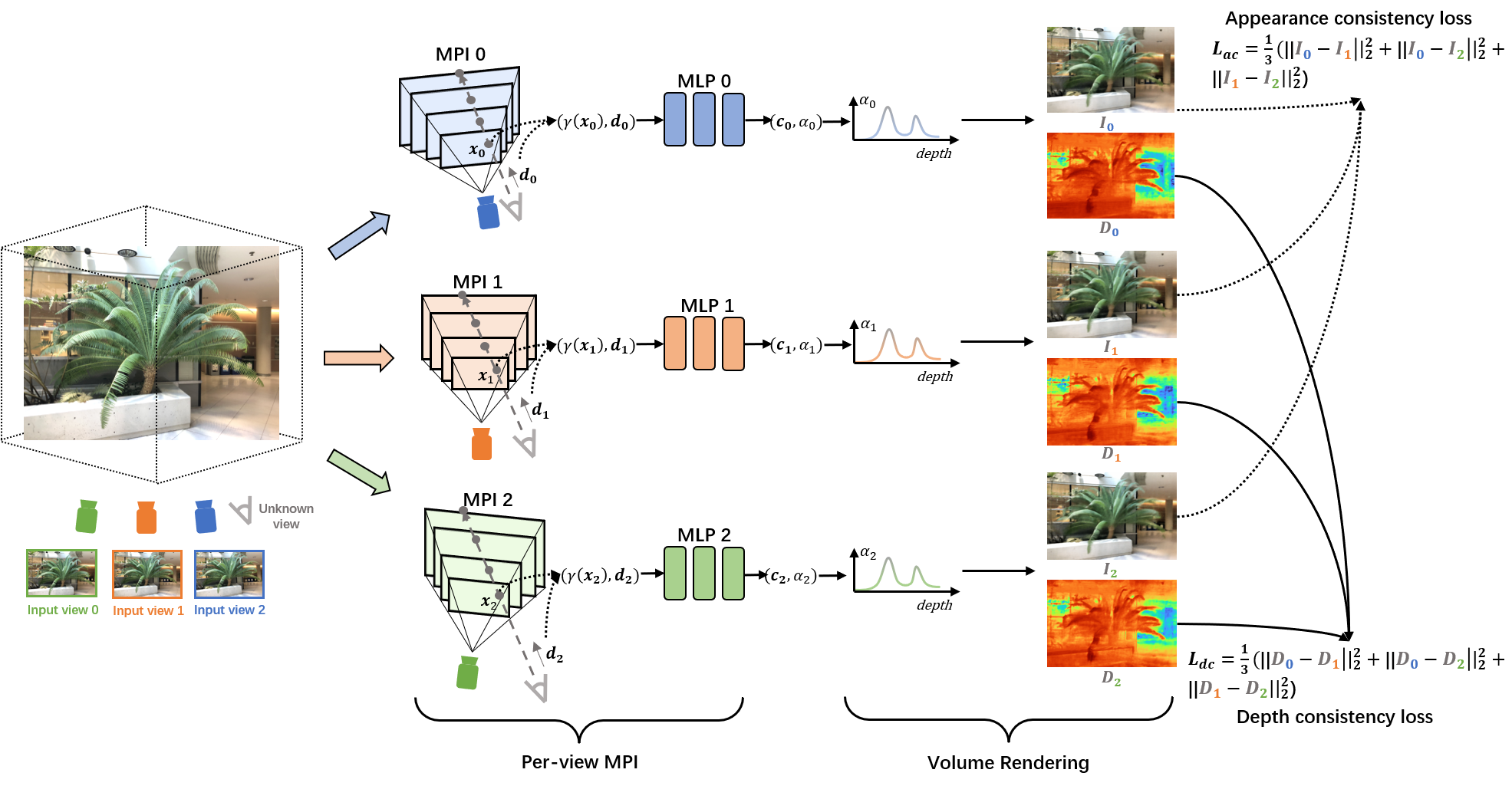}
  \caption{\textbf{Overview of our proposed method.} Given sparse input view images, we treat every input view as the reference view and construct their corresponding MPI respectively, where each MPI is parameterized by individual MLP (see Sec.~\ref{sec:representation} for details). Since the novel view image can be rendered by any MPI and deserve to have the same colors and depths, we propose the appearance and depth consistency loss to fully utilize cross-view multiplane consistency (see Sec.~\ref{sec:loss} for details).}
  \label{fig2: overview}
}

\abstract{Neural Radiance Field (NeRF) has shown impressive results in novel view synthesis, particularly in Virtual Reality (VR) and Augmented Reality (AR), thanks to its ability to represent scenes continuously. However, when just a few input view images are available, NeRF tends to overfit the given views and thus make the estimated depths of pixels share almost the same value. Unlike previous methods that conduct regularization by introducing complex priors or additional supervisions, we propose a simple yet effective method that explicitly builds depth-aware consistency across input views to tackle this challenge. Our key insight is that by forcing the same spatial points to be sampled repeatedly in different input views, we are able to strengthen the interactions between views and therefore alleviate the overfitting problem. To achieve this, we build the neural networks on layered representations (\textit{i.e.}, multiplane images), and the sampling point can thus be resampled on multiple discrete planes. Furthermore, to regularize the unseen target views, we constrain the rendered colors and depths from different input views to be the same. Although simple, extensive experiments demonstrate that our proposed method can achieve better synthesis quality over state-of-the-art methods.%
} % end of abstract

\CCScatlist{
  \CCScatTwelve{Neural Radiance Fields}{Few-shot view synthesis}{Multiplane Images}{Cross-view consistency}
}

\begin{document}

%% The ``\maketitle'' command must be the first command after the
%% ``\begin{document}'' command. It prepares and prints the title block.

%% the only exception to this rule is the \firstsection command
\firstsection{Introduction}

\maketitle

%% \section{Introduction} %for journal use above \firstsection{..} instead
As a fundamental task in computer vision and computer graphics, novel view synthesis aims at rendering novel view images from given several posed input view images \cite{buehler2001unstructured,debevec1996modeling}. 
% Traditional methods mainly rely on methods such as {image-based-rendering (IBR) to construct an explicit 3-dimensional model to render novel views} {[x]}.
Recently, Neural Radiance Field (NeRF)~\cite{mildenhall2021nerf} has gained increasing popularity due to its powerful ability in continuous scene representation and its superior performance of novel view synthesis.

% However, the success of NeRF and its following works depend on the number of input views to a large extent. When just a few input views are given, NeRF tends to overfit to input views and fails to model the scene geometry correctly, where the estimated depths of pixels in novel views share no difference. We refer to this behaviour as "depth vanishment" because it makes NeRF degenerate to a 2D plane model and the depth information of a 3D scene just disappear. Such a behaviour restrict the application of NeRF in many real scenarios, especially for those without enough input views.

\begin{figure}[t]
\begin{center}
\includegraphics[width=1\linewidth]{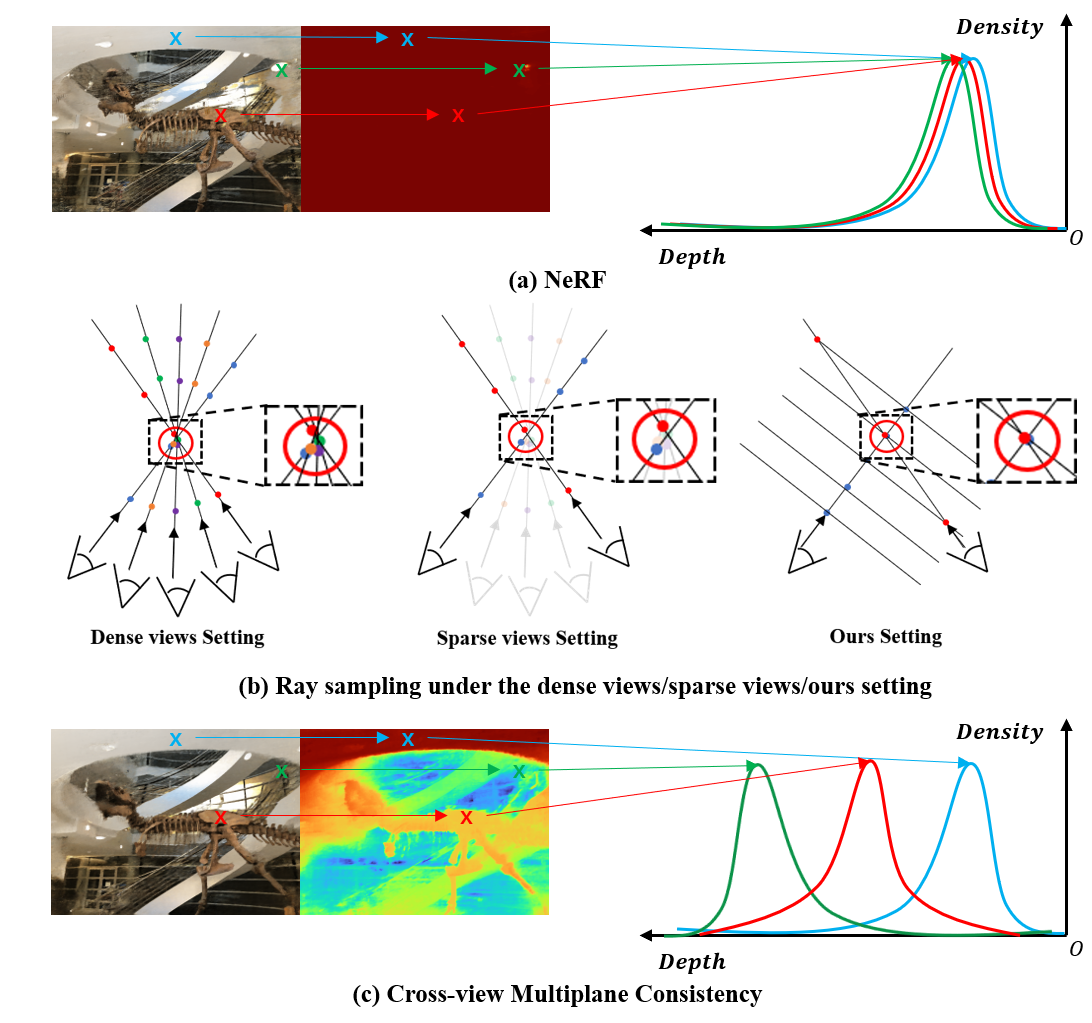}
\end{center}
\caption{Given a few input views (\textit{e.g.}, 3 input views), (a) NeRF tends to overfit to input views and results in a dramatic performance drop, where the estimated depths of pixels share almost the same value. (b) Our key insight is to ensure the same spatial points can be sampled repeatedly in different input views. (c) Our proposed method can achieve smooth depth estimation by introducing cross-view multiplane consistency, resulting in better synthesis quality.}
\label{fig1: qualitative comparison}
\end{figure}

However, the success of NeRF and its variants depends on the number of input views to a large extent~\cite{jain2021putting}. As shown in Fig.~\ref{fig1: qualitative comparison}(a), when just a few input views are given, NeRF tends to overfit input views, resulting in the estimated depths of pixels sharing almost the same value~\cite{jain2021putting,zhang2020nerf++}. In principle, this overfitting problem could be alleviated by incorporating priors of different scenes into the neural network~\cite{chen2021mvsnerf,chibane2021stereo,liu2022neural,trevithick2021grf,wang2022generalizable,wang2021ibrnet,yu2021pixelnerf}. However, these methods require expensive pre-training cost and the pre-trained scenes usually exist domain gap for the target scene~\cite{niemeyer2022regnerf}.

More recently, remarkable progress has also been made toward alleviating the overfitting problem by introducing external supervisions~\cite{deng2022depth,roessle2022dense,truong2022sparf}, pseudo views~\cite{ahn2022panerf,chen2022geoaug,kwak2023geconerf,truong2022sparf,xu2022sinnerf}, or physical priors~\cite{jain2021putting,kim2022infonerf,niemeyer2022regnerf}.
For example, Jain et al.~\cite{jain2021putting} introduced semantic consistency between various views to encourage realistic renderings. Niemeyer et al.~\cite{niemeyer2022regnerf} regularized the geometry and appearance of patches for each unseen view. Although effective, the aforementioned methods either ignore the consistency across multiple views~\cite{niemeyer2022regnerf,kim2022infonerf} or impose the cross-view consistency solely on the image level~\cite{jain2021putting}, thereby limiting the performance.

To tackle this challenge, we make an assumption: due to fewer input views, the sampling point in each ray would rarely be used to render other views, therefore the neural networks tend to memorize colors of each input view instead of learning the underlying geometry~\cite{zhang2021understanding,arpit2017closer}. To validate this assumption, we propose Cross-view Multiplane Consistency (CMC), in which we force the sampling points to remain identical when rendering different views, as demonstrated in Fig.~\ref{fig1: qualitative comparison}(b). In this way, the sampling points are able to be rendered to different-view images, resulting in depth-aware consistency across views. More specifically, for each input view, we build individual layered representations (\textit{i.e.,}, Multiplane Images) by regarding the input view as the reference view of the Multiplane Images (MPI)~\cite{zhou2018stereo}. Therefore, based on the discrete multiplane representation, all sampling points are forced to be distributed on the same fixed planes.

Given the multiplane representation for each input view, we aim at imposing cross-view consistency on multiplanes during the optimization. We recognize this in two aspects: 1) for the input views whose ground-truth images are available, we optimize each MPI using a reconstruction loss that minimizes the difference between the rendered input view images and the ground-truth input view images. 2) for the unseen views that lack ground-truth images, we leverage the underlying consistency: the colors and depths that are rendered from different input views (\textit{i.e.,}, different MPIs) should maintain the same. As a result, we achieve cross-view multiplanes consistency.

% Specifiaclly, as shown in Fig.~\ref{fig2: overview}, for each MPI we first use an independent neural network (i.e., one MLP) to represent it. Then, these MLPs are optimized by a reconstruction loss where the differences between the rendered input view images and the ground-truth input view images are minimized. Moreover, to make a better leverage of underlying consistency across unknown novel views, we further propose the appearance and depth consistency loss based on the fact that the rendered color and depth of novel view images by different MPIs should maintain the same. With these simple constraints, cross-view consistency is fully utilized and thus an elegant and better few-shot novel view synthesis can be achieved.

 We verify our assumptions and proposals on the common \textit{LLFF}~\cite{mildenhall2019local} and \textit{Shiny}~\cite{wizadwongsa2021nex} dataset, where the overfitting problem can be well overcame with a promising improvement in the qualities of synthetic novel views.

The main contributions of this paper can be summarized as follows:
\begin{itemize}[itemsep = 0pt,topsep = 2pt]
\item We propose to force the sampling points to be the same when rendering different views, which alleviates the overfitting problem of few-shot novel view synthesis.
\item To achieve cross-view multiplane consistency, in addition to reconstruction loss for input views, we propose to impose appearance and depth consistency to the unseen views. 
\item We provide an explanation for the overfitting problem and then give the intuition behind our proposed CMC.
\item Our proposed method achieves state-of-the-art performance on various widely adopted datasets.
\end{itemize}

\section{Related Work}
\subsection{Novel View Synthesis}
As a long-standing problem in computer vision and computer graphics, novel view synthesis has been studied for decades with methods based on image-based rendering \cite{buehler2001unstructured,chaurasia2013depth,debevec1996modeling,sinha2009piecewise}, light fields \cite{levoy1996light,mildenhall2019local,srinivasan2017learning,wood2000surface}, point clouds \cite{le2020novel,song2020deep,wiles2020synsin,you2022learning} and learning-based representation \cite{flynn2019deepview,flynn2016deepstereo,riegler2021stable,zhou2016view}. Recently people have witnessed an increasing popularity for Neural Radiance Field (NeRF)~\cite{mildenhall2021nerf} due to its remarkable performance for novel view synthesis. Given several 2D input view images of a static scene, NeRF can render photorealistic novel view images through coordinate-based implicit neural representation. It has been extended to several different tasks, such as dynamic scenes representation \cite{park2021nerfies,pumarola2021d}, fast training and rendering \cite{fridovich2022plenoxels,muller2022instant,wizadwongsa2021nex,yu2021plenoctrees}, stylization \cite{huang2022stylizednerf,mu20223d,
wang2022nerf}, generalizable scenes representation \cite{chen2021mvsnerf,liu2022neural,wang2021ibrnet,xu2022point} etc. Though NeRF achieved great synthesis quality, it depends on dense input view images, which would be not suitable for many practical applications. As a result, in this paper, we focus our attention on view synthesis with sparse input views, \textit{e.g.}, few-shot novel view synthesis.

\subsection{Few-shot NeRF}
When only a few input view images with big disparities are available, NeRF easily overfits these input views, as shown in Fig.~\ref{fig1: qualitative comparison}(a). Some generalizable neural fields \cite{chen2021mvsnerf,chibane2021stereo,wang2021ibrnet,yu2021pixelnerf} could avoid this problem by using large-scale cross-scenes datasets to learn scenes priors, while the performance will degrade significantly when there is a large domain gap between the test scenes and the training dataset. \cite{deng2022depth,roessle2022dense,truong2022sparf} proposed to overcome the overfitting tendency of the few-shot setting in a per-scene optimization manner with additional supervision signals, such as sparse depth estimated by Structure-from-Motion~\cite{schonberger2016structure} or pixel correspondence estimated by \cite{truong2021learning}. To increase the number of training views available, \cite{ahn2022panerf,chen2022geoaug,kwak2023geconerf,xu2022sinnerf} proposed to use depth-warping to generate novel view images as pseudo labels. \cite{jain2021putting,kim2022infonerf,niemeyer2022regnerf} made use of physical priors to regularize the scene geometry without any additional supervision signals. {Recently, FreeNeRF~\cite{yang2023freenerf} mitigated the overfitting problem from the pespective of frequency, where a novel frequency annealing strategy on positional encoding was proposed. SimpleNeRF~\cite{somraj2023simplenerf} instead leveraged augmented models for better and stable few-shot view synthesis. MixNeRF~\cite{seo2023mixnerf} modeled rays as mixtures of Laplacianssians, followed by FlipNeRF~\cite{seo2023flipnerf} which 
used flipped reflection rays as additional training sources.}

Though these methods would achieve promising results, they either heavily rely on pre-trained neural networks that are usually expensive~\cite{roessle2022dense}, or only take advantage of physical priors as regularization terms on seen/unseen views independently, without cross-view interactions~\cite{kim2022infonerf}. Instead, in this paper, we propose to make full use of cross-view consistency to achieve the few-shot novel view synthesis.

\subsection{Multiplane Images}
MPI was first proposed by \cite{zhou2018stereo} to expand the small baselines of stereo images. Then \cite{mildenhall2019local} extended MPI to view synthesis by constructing local MPIs and blending different MPIs to render novel views. To achieve a fast generation of MPI, DeepView was proposed by \cite{flynn2019deepview} through the leverage of learned gradient descent. To model the time-dependent effects of scenes shot at different times, DeepMPI was introduced by \cite{li2020crowdsampling} in an unsupervised manner. 
\cite{han2022single,li2021mine,tucker2020single} further proposed to use MPI to realize single-view synthesis. Recently \cite{wizadwongsa2021nex} has been proposed to model view-dependent effects and to realize real-time rendering. Then \cite{zhao2022generative} proposed to take advantage of MPIs to make a 2D GAN 3D-aware. In this paper, we first apply MPI to few-shot view synthesis, where every input view is treated as the reference view respectively. To enhance the interactions across different views, we propose two new loss functions, \textit{i.e.}, the appearance and depth consistency loss, based on the fact that the rendered colors and depths of the target view by different MPIs should be the same.

\section{Preliminaries} \label{Preliminaries}

Our method is built upon Neural Radiance Field (NeRF)~\cite{mildenhall2021nerf} and Multiplane Images (MPI)~\cite{zhou2018stereo}. We elaborate on them in this section.

\subsection{Neural Radiance Field}
NeRF~\cite{mildenhall2021nerf} has emerged as a powerful tool for continuous scene representation by encoding scene properties into a neural network $F_{\theta}$, which is usually parameterized by one Multilayer Perceptron (MLP). Input the 3D coordinate $\mathbf{x}=(x,y,z)$ of a spatial point and one viewing direction $\mathbf{d}=(d_x,d_y,d_z)$, NeRF outputs the corresponding color $\mathbf{c}$ and volume density $\sigma$, which is denoted as:
\begin{equation}
  \centering \label{equ_3.1}
    \mathbf{c},\sigma = F_{\theta}(\gamma(\mathbf{x}),\gamma(\mathbf{d})),
\end{equation}
where $\gamma$ is the position encoding operation \cite{mildenhall2021nerf} that aim to recovering high-frequency detail textures.

Given several input view images and their camera parameters, a pixel can be rendered by casting a ray $\mathbf{r}(t)=\mathbf{o}+t\mathbf{d}$ from the camera origin $\mathbf{o}$ towards the pixel along direction $\mathbf{d}$. Specifically, assuming $t\in [t_n,t_f]$, the estimated color $\mathbf{C}(\mathbf{r})$ of this pixel is formulated as follows:
\begin{equation}
  \centering \label{equ_3.2}
    \mathbf{C}(\mathbf{r})=\int_{t_n}^{t_f} T(t) \sigma(\mathbf{r}(t)) \mathbf{c}(\mathbf{r}(t), \mathbf{d}) d t,
\end{equation}
where $T(t)=\exp \left(-\int_{t_n}^{t} \sigma(\mathbf{r}(s)) d s\right)$, $\sigma$ and $\mathbf{c}$ are obtained by Eq. \ref{equ_3.1}. NeRF is optimized by minimizing the following loss function:
\begin{equation}
  \centering \label{equ_3.3}
    \mathcal{L}_{\text{MSE}}=\frac{1}{|\mathcal{R}|} \sum_{\mathbf{r} \in \mathcal{R}}\|\mathbf{C}(\mathbf{r})-\mathbf{C}_{gt}\|_2^2,
\end{equation}
where $\mathcal{R}$ is a set of sampling rays, $\mathbf{C}(\mathbf{r})$ is obtained by Eq. \ref{equ_3.2} and $\mathbf{C}_{gt}$ represents the ground-truth color.

\subsection{Multiplane Images}
As a layered scene representation, MPI \cite{zhou2018stereo} is constructed by a set of frontop-parallel planes with respect to a reference view, where all planes are fixed at specific depths that are distributed equally in the depth space. Considering one MPI with $D$ planes $(\mathbf{c}_{i},\alpha_{i})_{i=1}^{D}$, the ${i}$-th plane at depth $z_i$ can be viewed as a $4$-channel RGBA image that contains the color $\mathbf{c}_{i}$ and visibility $\alpha_{i}$.

To render a target view based on the MPI of the reference view, each plane of the MPI is warped to the target view $(\mathbf{c}_{i}^{\prime},\alpha_{i}^{\prime})_{i=1}^{D}$ using inverse homography warping, followed by an alpha-composition operation \cite{han2022single,li2021mine,tucker2020single,zhou2018stereo}. The rendered image $\mathbf{I}_t$ and depth map ${Z}_t$ of target view are denoted as follows:
\begin{equation}
  \centering \label{equ_3.4}
    \mathbf{I}_t=\sum_{i=1}^D\left(\mathbf{c}_i^{\prime} \alpha_i^{\prime} \prod_{j=1}^{i-1}\left(1-\alpha_j^{\prime}\right)\right)
\end{equation}

\begin{equation}
  \centering \label{equ_3.5}
    {Z}_t=\sum_{i=1}^D\left(z_i \alpha_i^{\prime} \prod_{j=1}^{i-1}\left(1-\alpha_j^{\prime}\right)\right).
\end{equation}

We build our model on MPI. Therefore, the sampling point can be resampled on multiple discrete planes.

\section{Cross-view Multiplane Consistency}
\paragraph{Motivation.}
As shown in Fig.~\ref{fig1: qualitative comparison}(a), NeRF suffers from significant performance degradation when the number of input views is reduced, which also leads to the estimated depths of pixels sharing almost the same value~\cite{jain2021putting,zhang2020nerf++}. To tackle this problem, we assume that one plausible reason is that the sampling point in each ray would rarely be used to render other views due to fewer input views. Therefore, it is easier for the neural networks to memorize each input view images~\cite{zhang2021understanding,arpit2017closer}, rather than learning the underlying geometry. Motivated by this, our key insight is to explicitly build depth-aware consistency across different views.

\paragraph{Method Overview.}
As shown in Fig.~\ref{fig2: overview}, to ensure that the sampling points are the same when rendering different views, we build individual layered representation (\textit{i.e.}, Multiplane Images) $F_{\theta}^i$ for each input view $i$ by utilizing the input view $i$ as the reference view of the Multiplane Images (MPI)~\cite{zhou2018stereo}. Therefore, all sampling points are distributed on the same fixed planes. Inspired by previous works~\cite{wizadwongsa2021nex,li2021mine}, each MPI is presented by a multilayer perceptron (MLP) $F_{\theta}^i$, which outputs the color and visibility for each plane.

To optimize $F_{\theta}^i$, for the input views, we directly minimize the difference between rendered images and the ground-truth ones through a reconstruction loss. While there is no ground-truth image for the unseen views, we introduce an intuition that the colors and depth rendered by different input views should be the same. Specifically, we minimize the difference in the estimated colors and depths that are obtained by different MPIs.

\subsection{Multiplane Representation for Input Views}
\label{sec:representation}

As shown in Fig.~\ref{fig2: overview}, given several sparse input view images $\{\mathbf{I}_{in}^{i}\}_{i=0}^{N-1}\in \mathit{R}^{H\times W\times 3}$ and their corresponding camera extrinsics $[\mathbf{R}_{in}^{i}|\mathbf{t}_{in}^{i}]_{i=0}^{N-1} \in SE(3)$ of a static scene, our goal is to render novel view images photorealisticly, where $H$ and $W$ are the image height and width, $N$ is the number of input views available, $\mathbf{R} \in \mathit{R}^{3 \times 3}$ and $\mathbf{t} \in \mathit{R}^{3 \times 1}$ represent the rotation matrix and translation vector.

As described in motivation, in this paper we use MPIs to represent the scene. Different from most MPI-based methods that randomly choose one input view as the reference view and the left input views as the target views \cite{han2022single,li2021mine,tucker2020single,wizadwongsa2021nex,zhou2018stereo}, we propose to treat every input view as the reference view respectively and construct their corresponding MPIs $\{\mathbf{M}_i\}_{i=0}^{N-1}$ (\textit{i.e.}, per-view MPI), for the purpose of building depth-aware consistency across different input views. We adopt MLPs to present the MPIs following previous works~\cite{wizadwongsa2021nex,li2021mine}.

Specifically, considering the camera parameter $[\mathbf{R}_{t}|\mathbf{t}_{t}]$ of one target view and the $i$-th MPI $\mathbf{M}_i$ corresponding to $\mathbf{I}_{in}^i$ that has $D$ planes. When a ray $\mathbf{r}$ is cast from the camera origin $\mathbf{o}$ of the target view through one pixel at its image plane whose coordinate is $(u_t,v_t)$ along direction $\mathbf{d}_i^{(u_t,v_t)}$, it will have $D$ intersections with the $D$ planes of $\mathbf{M}_i$, which are denoted as $\{\mathbf{x}_i^k = (u_i^k,v_i^k,z_i^k)\}_{k=0}^{D-1}$, where $(u_i^k,v_i^k)$ is the pixel coordinate of the $k$-th intersection and $z_i^k$ represents the depth that plane $k$ is placed. The pixel coordinate of each intersection can be computed by the inverse homography warping operation \cite{han2022single,li2021mine,tucker2020single,zhou2018stereo,wizadwongsa2021nex}, which is formulated as follows:
\begin{equation}
  \centering \label{equ_4.1}
    \left[\begin{array}{c}
u_i^k \\
v_i^k \\
1
\end{array}\right] \sim \mathbf{K}_{in}^i\left(\mathbf{R}^{\prime}-\frac{\mathbf{t}^{\prime} \mathbf{n}^{\top}}{z_i^k}\right) \mathbf{K}_{t}^{-1}\left[\begin{array}{c}
u_t \\
v_t \\
1
\end{array}\right],
\end{equation}
where $\mathbf{K}_{in}^i \in \mathit{R}^{3 \times 3}$ and $\mathbf{K}_{t} \in \mathit{R}^{3 \times 3}$ are the camera intrinsics for the input view $\mathbf{I}_{in}^i$ and target view respectively, $\mathbf{n}=[0,0,1]^{\top}$ is the normal vector of the $k$-th plane, $\mathbf{R}^{\prime}$ and $\mathbf{t}^{\prime}$ are the relative camera extrinsic from the target view to the input view, which is computed as follows:
\begin{equation}
  \centering \label{equ_4.2}
   \begin{bmatrix} \mathbf{R}^{\prime}_{3\times3}  & \mathbf{t}^{\prime}_{3\times1}  \\ \mathbf{0}_{1\times3}  & 1 \\ \end{bmatrix} = \begin{bmatrix} \mathbf{R}_{t} & \mathbf{t}_{t} \\ \mathbf{0}_{1\times3} & 1 \\ \end{bmatrix}^{-1} \begin{bmatrix} \mathbf{R}_{in}^{i} & \mathbf{t}_{in}^{i} \\ \mathbf{0}_{1\times3}  & 1 \\ \end{bmatrix}.
\end{equation}

With the computed coordinate $\mathbf{x}_i^k$ of each intersection along the ray $\mathbf{r}$ whose direction is $\mathbf{d}_i^{(u_t,v_t)}$, both $\mathbf{x}_i^k$ and $\mathbf{d}_i^{(u_t,v_t)}$ are fed into the MLP $F_{\theta}^i$ to estimate its color $\mathbf{c}_i^k$ and visibility $\alpha_i^k$ as shown in Fig.~\ref{fig2: overview}, which is denoted as:
\begin{equation}
  \centering \label{equ_4.3}
  \mathbf{c}_i^k,\alpha_i^k = F_{\theta}^i(\gamma(\mathbf{x}_i^k),\gamma(\mathbf{d}_i^{(u_t,v_t)})),
\end{equation}
where $\gamma$ is the position encoding operation \cite{mildenhall2021nerf} that is formulated as follows:
\begin{equation}
\centering \label{equ_4.4}
  \gamma(\mathbf{x})=(\sin(2^0\mathbf{x}),\cos(2^0\mathbf{x}),\cdots, \sin(2^{L-1}\mathbf{x}),\cos(2^{L-1}\mathbf{x})),
\end{equation}
$L$ is the hand-crafted hyperparameter. Then the color $\mathbf{C}_{i}(\mathbf{r})$ and depth $Z_{i}(\mathbf{r})$ of the pixel $(u_t,v_t)$ in the target view can be rendered based on {volume rendering} by the $i$-th MPI.

\subsection{Cross-view Consistency on Multplanes}
\label{sec:loss}

\paragraph{Reconstruction Loss for Input Views.}
Given the rendered color $\mathbf{C}_{i}(\mathbf{r})$, if the target view is one of the input views, then the reconstruction loss (Eq.~\ref{equ_3.3}) that minimizes the difference from $\mathbf{C}_{i}(\mathbf{r})$ to the ground truth color $\mathbf{C}_{gt}$ is adopted, which is denoted as follows:
\begin{equation}
\centering \label{equ_4.5}
  \mathcal{L}_{\text{MSE}}=\frac{1}{|{N}|}\frac{1}{|\mathcal{R}|}\sum_{i=0}^{N-1} \sum_{\mathbf{r} \in \mathcal{R}}\|\mathbf{C}_{i}(\mathbf{r})-\mathbf{C}_{gt}\|_2^2,
\end{equation}
where $\mathcal{R}$ is a set of sampling rays. In Sec.~\ref{ablation: single MPI}, we verify that with our multiplane representation, the reconstruction loss alone can overcome the overfitting problem well.

\paragraph{Appearance and Depth Consistency Loss for Unseen Views.}
The above reconstruction loss utilizes consistency across known input views by forcing the spatial points to be sampled on the same planes. To obtain depth-aware consistency across views, we propose the appearance and depth consistency loss across unseen novel views.

Specifically, when the target view is a novel view that has no ground truth color, it still can be rendered by any MPI and deserve to have the same color and depth map, as shown in Fig.~\ref{fig2: overview}. Based on such an observation, we propose the following loss functions:  
\begin{equation}
\centering \label{equ_4.6}
  \mathcal{L}_{\mathrm{ac}}=\frac{2}{|{N(N-1)}|}\frac{1}{|\mathcal{R}|}\sum_{i=0}^{N-1} \sum_{j=i+1}^{N-1} \sum_{\mathbf{r} \in \mathcal{R}}\|\mathbf{C}_{i}(\mathbf{r})-\mathbf{C}_{j}(\mathbf{r})\|_2^2,
\end{equation}
\begin{equation}
\centering \label{equ_4.7}
  \mathcal{L}_{\mathrm{dc}}=\frac{2}{|{N(N-1)}|}\frac{1}{|\mathcal{R}|}\sum_{i=0}^{N-1} \sum_{j=i+1}^{N-1} \sum_{\mathbf{r} \in \mathcal{R}}\|{Z}_{i}(\mathbf{r})-{Z}_{j}(\mathbf{r})\|_2^2,
\end{equation}
where $\mathbf{C}_{i}$/${Z}_{i}$ and $\mathbf{C}_{j}$/${Z}_{j}$ represents the rendered colors and depths by the $i$-th MPI and $j$-th MPI respectively.

As a result, the whole loss function of our proposed method can be expressed as follows:
\begin{equation}
\centering \label{equ_4.8}
  \mathcal{L} = \mathcal{L}_{\text{MSE}} + \lambda_{ac}\mathcal{L}_{\mathrm{ac}} + \lambda_{dc}\mathcal{L}_{\mathrm{dc}},
\end{equation}
where $\lambda_{ac}$ and $\lambda_{dc}$ are hyperparametes that balance the weights of $\mathcal{L}_{\mathrm{ac}}$ and $\mathcal{L}_{\mathrm{dc}}$.

% Table generated by Excel2LaTeX from sheet 'Sheet1'
\begin{table*}[htbp]
	\centering
	\caption{Quantitative comparisons on 8 scenes of the \textit{Shiny} dataset.}
	\resizebox{\textwidth}{!}
	{ 
	\begin{tabular}{c|ccc|ccc|ccc|ccc}
		\hline
        \hline
		Method & \multicolumn{3}{c|}{NeRF \cite{mildenhall2021nerf}} & \multicolumn{3}{c|}{DietNeRF \cite{jain2021putting}} & \multicolumn{3}{c|}{InfoNeRF \cite{kim2022infonerf}} & \multicolumn{3}{c}{Ours} \bigstrut\\
		\hline
		Scene & PSNR$\uparrow$  & SSIM$\uparrow$  & LPIPS$\downarrow$ & PSNR$\uparrow$  & SSIM$\uparrow$  & LPIPS$\downarrow$ & PSNR$\uparrow$  & SSIM$\uparrow$  & LPIPS$\downarrow$ & PSNR$\uparrow$  & SSIM$\uparrow$  & LPIPS$\downarrow$ \bigstrut\\
		\hline
		Cake  & 15.98 & 0.514 & 0.576 & 18.04 & 0.556 & 0.543 & 14.71 & 0.469 & 0.653 & {17.08} & {0.564} & {0.496} \bigstrut[t]\\
		Crest & 11.50  & 0.152 & 0.729 & 9.74  & 0.105 & 0.733 & 12.28 & 0.181 & 0.736 & {14.54} & {0.268} & {0.564} \\
		Food  & 12.65 & 0.296 & 0.657 & 10.30  & 0.190  & 0.736 & 13.25 & 0.328 & 0.679 & {16.00} & {0.425} & {0.502} \\
		Giants & 12.39 & 0.218 & 0.730  & 12.54 & 0.216 & 0.733 & 6.32  & 0.010  & 0.776 & {13.42} & {0.299} & {0.651} \\
		Pasta & 13.95 & 0.370  & 0.550  & 13.96 & 0.373 & 0.545 & 13.84 & 0.353 & 0.632 & {14.89} & {0.389} & {0.523} \\
		Room  & 21.19 & 0.710  & 0.454 & 20.01 & 0.669 & 0.483 & 18.99 & 0.578 & 0.638 & {22.59} & {0.750} & {0.378} \\
		Seasoning & 12.27 & 0.358 & 0.684 & 12.05 & 0.347 & 0.682 & 12.62 & 0.384 & 0.684 & {13.05} & {0.447} & {0.605} \\
		Tools & 15.04 & 0.580  & 0.500   & 8.35  & 0.276 & 0.717 & 10.89 & 0.358 & 0.65  & {16.23} & {0.598}  & {0.411} \bigstrut[b]\\
		\hline
		Average & 14.37 & 0.399 & 0.610  & 13.12 & 0.341 & 0.646 & 12.86 & 0.332 & 0.681 & {\textbf{15.98}} & {\textbf{0.468}} & {\textbf{0.516}} \bigstrut\\
		\hline
        \hline
	\end{tabular}%
}
	\label{Quantitative comparisons on 8 scenes of the shiny dataset}%
\end{table*}%

{\subsection{Weighted Rendering}
To render a target view from multiple MPIs, based on the assumption that the closest MPI to the target view should have a greater impact on its rendering process, we adopt a weighted rendering strategy. Specifically,  the final output $\mathbf{C}(\mathbf{r})$ is obtained by calculating a weighted average of the rendering colors from different MPIs, which is denoted as follows:
\begin{equation}
\centering \label{equ_weighted_rendering}
  \mathbf{C}(\mathbf{r}) = \sum_{i=0}^{N-1}w_i\cdot\mathbf{C}_{i}(\mathbf{r}),
\end{equation}
where $\mathbf{C}_{i}(\mathbf{r})$ is the color rendered by the $i$-th MPI, $w_i$ is the weight  calculated according to the distance $\mu_i$ from the $i$-th MPI to the target view, which is formulated as follows:
\begin{equation}
\centering \label{equ_weight_calculation}
  w_i = \frac{\mu_i}{\sum_{j=0}^{N-1}\mu_j},\ \mu_i = ||\mathbf{o}_t - \mathbf{o}_i||_2^2,
\end{equation}
where $\mathbf{o}_t$ and $\mathbf{o}_i$ represent the camera origins of the target view and the $i$-th MPI respectively.

}

\subsection{Analysis on Cross-view Multiplane Consistency}
To demonstrate the effectiveness of our proposed method, we make an analysis of CMC in this section. To begin with, we propose an assumption for the overfitting problem of NeRF under the few-shot setting. Specifically, given sparse input views, as shown in Fig.~\ref{fig1: qualitative comparison}(b), a fact is that it is quite difficult for rays of different views to have the same sampling points due to the random uniform sampling strategy of NeRF \cite{mildenhall2021nerf}, which is denoted as follows:
\begin{equation}
\centering \label{equ_4.9}
t_i \sim \mathcal{U}\left[t_n+\frac{i-1}{M}\left(t_f-t_n\right), t_n+\frac{i}{M}\left(t_f-t_n\right)\right],
\end{equation}
where $t_n$ and $t_f$ are the near and far bounds, $M$ is the number of sampling points along the ray and $t_i$ is the $i$-th sampling points. As a result, our assumption is that the sampling points in each ray would only take part in the rendering process of pixels corresponding to this ray, while rarely being used to render other views. Thus, the optimization process of NeRF~\cite{mildenhall2021nerf} (Eq.~\ref{equ_3.2}) can be viewed as solving the following equation for each ray independently:
\begin{equation}
\centering \label{equ_4.10}
\mathbf{C}_{gt}=\sum_{i=1}^M T(\sigma_i) f(\sigma_i) \mathbf{c}_i, 
\end{equation}
where $T$ and $f$ are both functions of $\sigma_i$, $\mathbf{C}_{gt}$ is the ground-truth pixel of ray $\mathbf{r}$, $\sigma_i$ and $\mathbf{c}_i$ are unknowns to be estimated. Obviously, we have $2M$ unknowns while only one equation, which means that infinite solutions exist for this problem. Considering the memorization nature of neural networks~\cite{zhang2021understanding,arpit2017closer} and Occam's Razor~\cite{blumer1987occam,rasmussen2000occam}, NeRF tends to converge to the simplest way to represent known input views, thus Eq.~\ref{equ_4.10} is assumed to solve the following sparse optimization problem:
\begin{equation}
\centering \label{equ_4.11}
\begin{aligned}
\mathbf{c}_i^{*},\sigma_i^{*} = \mathop{\arg\min}_{\mathbf{c}_i,\sigma_i} \{||&\sum_{i=1}^M T(\sigma_i) f(\sigma_i) \mathbf{c}_i - \mathbf{C}_{gt}||_2^{2} \\+ & \sum_{i=1}^M ||\mathbf{c}_i||_0 + \sum_{i=1}^M ||\sigma_i||_0\}
\end{aligned},
\end{equation}
whose solution is 
\begin{equation}
\label{equ_4.12}
\{\mathbf{c}_i^*,\sigma_i^*\}=\left\{
\begin{aligned}
&\{\mathbf{C}_{gt},1\}  , & i=0 \\
&\{\mathbf{0},0\}  , & i\geq1
\end{aligned}
\right.,
\end{equation}
which thus leads to the overfitting problem.

Based on the analysis above, to overcome such a problem, a direct way is to impose the same point to be sampled in rays of different views. Take two input views $I_0$ and $I_1$ as an example, whose camera origins are $\mathbf{o}_0$ and $\mathbf{o}_1$ respectively. For one sampling point $\mathbf{x}_0 = \mathbf{o}_0 + t_0\mathbf{d_0}$ along ray $\mathbf{d_0}$ of $I_0$, our goal is that $\mathbf{x}_0$ can also be sampled in $I_1$ along ray $\mathbf{d_1}$, thus guarantee that the same sampling point can take part in the rendering process of pixels in different views. Assuming that the sampling point in ray $\mathbf{d_1}$ is denoted as $\mathbf{x}_1 = \mathbf{o}_1 + t_1\mathbf{d_1}$, then the problem can be converted into the following formulation (\textit{i.e.}, find the optimal $t_1$ that can minimize the distances between $\mathbf{x}_0$ and $\mathbf{x}_1$):
\begin{equation}
\centering \label{equ_4.13}
\begin{aligned}
t_1^{*} = \mathop{\arg\min}_{t_1} & ||\mathbf{x}_1 - \mathbf{x}_0||_2^{2}, \, t_1 \in [t_n,t_f],
\end{aligned}
\end{equation}
where $\mathbf{x}_0$, $\mathbf{o}_1$ and $\mathbf{d_1}$ is known. {As a result, our goal is to find the optimal $t_1$ that can satisfy the following formulation:
\begin{equation} \label{eq 30}
\centering
\begin{aligned}
||\mathbf{x}_1 - \mathbf{x}_0||_2^{2} \leq \delta 
\end{aligned},
\end{equation}
where $\delta \to 0$.

Assuming that $\mathbf{o}_0 = \{o_0^x,o_0^y,o_0^z\}$, $\mathbf{d}_0 = \{d_0^x,d_0^y,d_0^z\}$, $\mathbf{o}_1 = \{o_1^x,o_1^y,o_1^z\}$ and $\mathbf{d}_1 = \{d_1^x,d_1^y,d_1^z\}$, then Eq.~\ref{eq 30} can be converted into the following formulation:
\begin{equation} 
\centering
\begin{aligned}
&||\{o_0^x+t_0d_0^x,o_0^y+t_0d_0^y,o_0^z+t_0d_0^z\} - \\ &\{o_1^x+t_1d_1^x,o_1^y+t_1d_1^y,o_1^z+t_1d_1^z\}||_2^{2} \leq \delta 
\end{aligned}.
\end{equation}

Simplifying the above formula, we obtain:
\begin{equation} 
\centering
\begin{aligned}
t_1^* = \Phi(t_0) 
\end{aligned},
\end{equation}
where 
\begin{equation} 
\centering
\begin{aligned}
\Phi(u) = ((o_0^x&+ud_0^x - o_1^x)^2 + (o_0^y+ud_0^y - o_1^y)^2 \\
&+ (o_0^z+ud_0^z - o_1^z)^2)^{1/2}
\end{aligned},
\end{equation}}
which means that when sampling points in view $I_0$ are known, then all sampling points in view $I_1$ should be deterministic. Fortunately, this is exactly the nature of multiplane images. When view $I_0$ is selected as the reference view to construct the MPI, all the sampling points of different views are deterministic and forced to be distributed on the same planes. As a result, consistency across different views can be well guaranteed. Experiments in Sec.~\ref{ablation: single MPI} also demonstrate the effectiveness of our analysis.

\section{Experiments}
We make a comparison with various state-of-the-art methods for few-shot novel view synthesis quantitatively and qualitatively. We also present a detailed analysis of the necessity of adopting per-view multiplane images and the appearance/depth consistency loss. See supplementary materials for demonstrations of Eq.~\ref{equ_4.11} and Eq.~\ref{equ_4.13}, ablation studies on the influence of different numbers of MPI planes, and more visualization results of novel view synthesis. {We only evaluate our proposed method on extremely sparse input views, \textit{i.e.}, 3 input views, as it is the most common case.}

\begin{figure}[t]
\begin{center}
\includegraphics[width=1\linewidth]{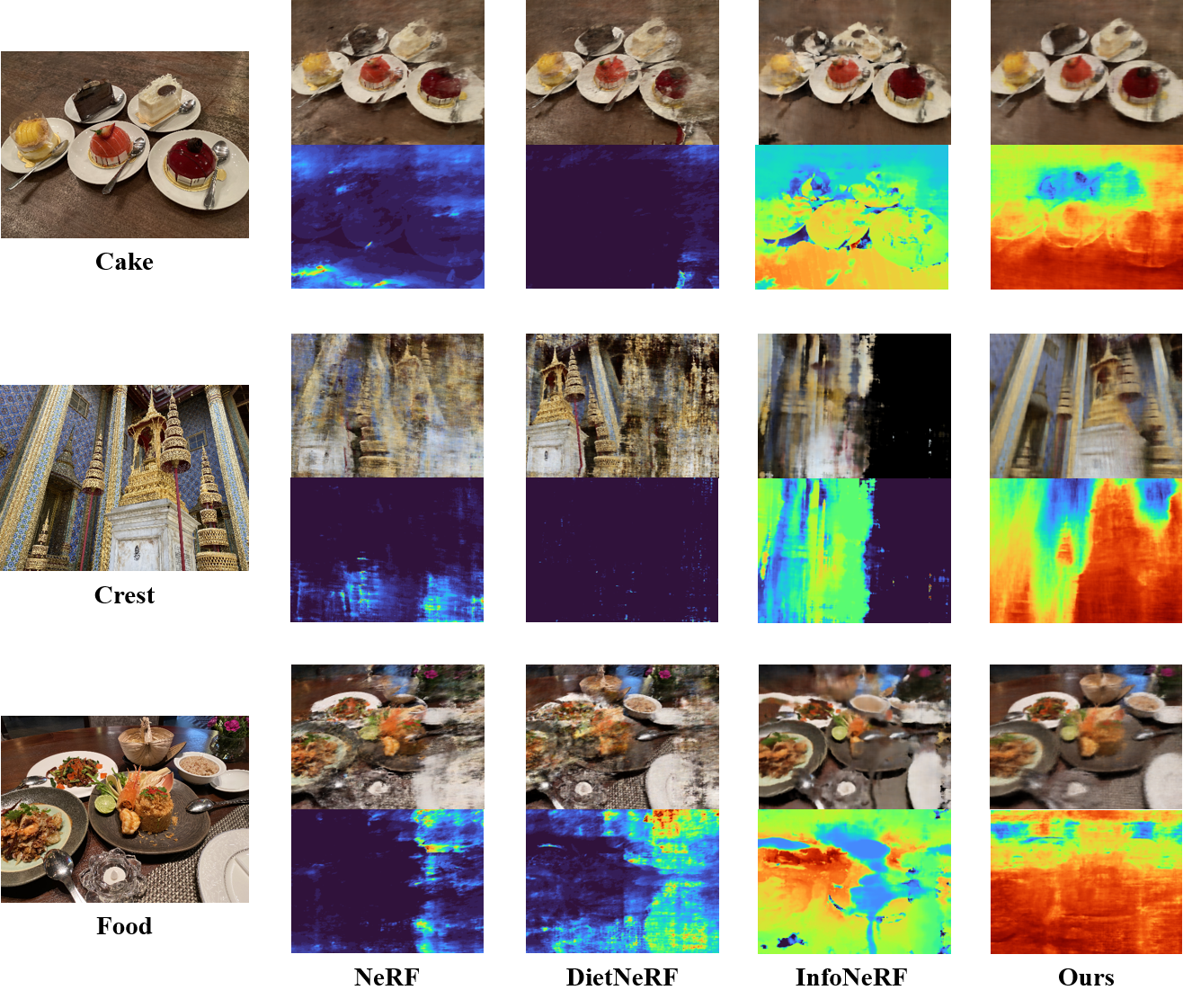}
\end{center}
\caption{Qualitative comparisons on the \textit{Shiny} dataset, where our proposed method can achieve better novel view synthesis and accurate geometry estimation (\textit{i.e.}, the depth map).}
\label{fig3: Qualitive comparisons on the shiny dataset.}
\end{figure}

\begin{figure*}[t]
	\begin{center}
		% \rule{0pt}{2in} \rule{.9\linewidth}{0pt}
		\includegraphics[width=\linewidth]{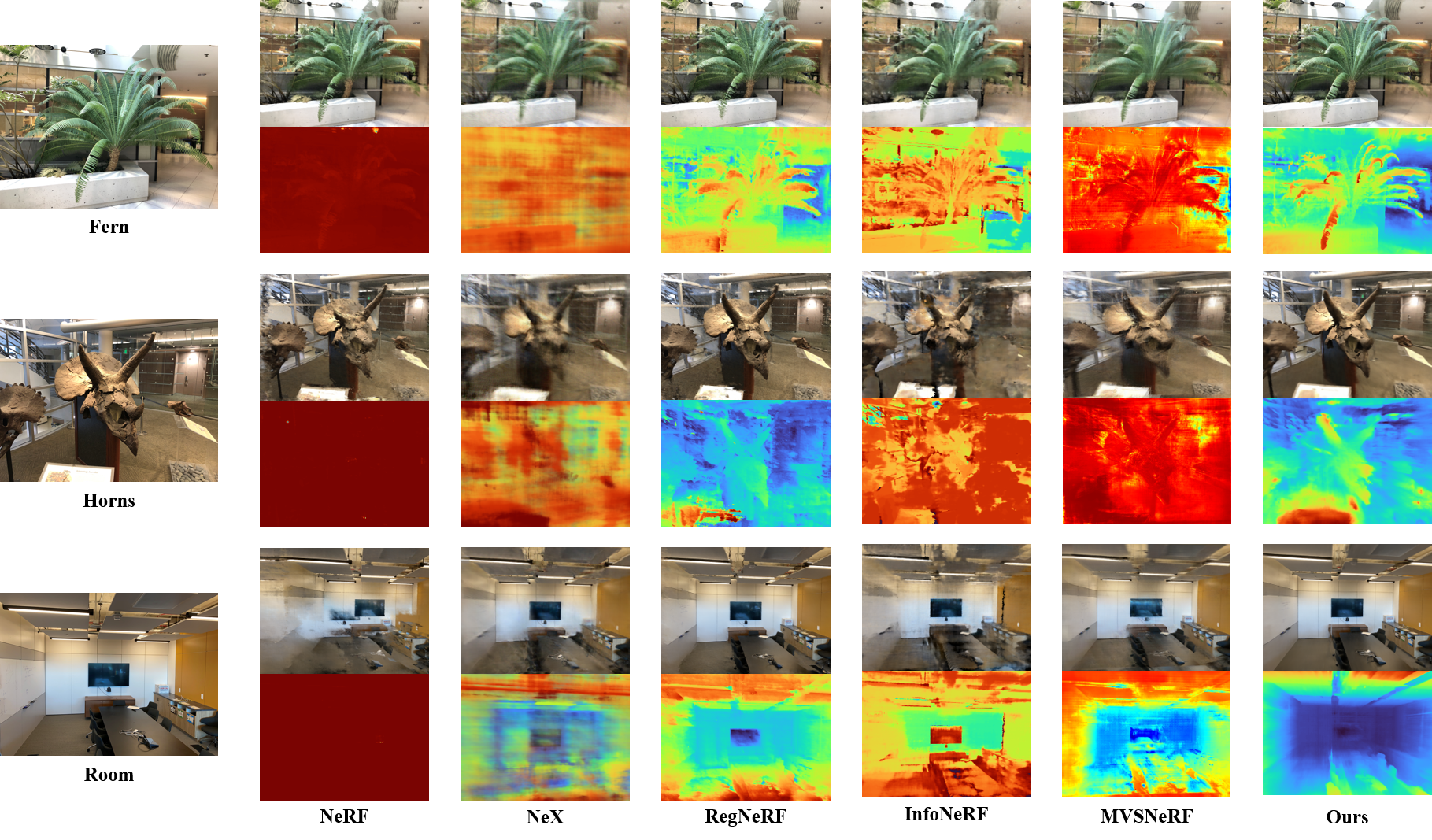}
	\end{center}
	\caption{Qualitative comparisons on the \textit{LLFF} dataset. Our proposed method can avoid the overfitting problem, where better novel view synthesis and more continuous depth estimation can be achieved.}
	\label{fig4: Qualitive comparisons on the llff dataset}
\end{figure*}

\subsection{Implementation Details}
\paragraph{Datasets.} 
We perform experiments on the \textit{LLFF} dataset~\cite{mildenhall2019local} and the \textit{Shiny}~\cite{wizadwongsa2021nex} dataset to validate the effectiveness of our proposed method.
Both of the two datasets contain 8 complex real-world scenes with big disparities, while the \textit{Shiny} dataset is more complicated because it has more view-dependent effects such as reflections and refraction. {We follow the experimental protocols provided by~\cite{niemeyer2022regnerf}, where the resolution of both input views and target views are 378 $\times$ 504.} {To make a fair comparison}, similar to previous methods, for each scene we choose every 8-th image as the held-out test set and then select 3 images evenly from the remaining images as the input views. {Notably, following~\cite{niemeyer2022regnerf}, in our experiment the sampled input views are distributed uniformly in the camera pose space, where the distances across different input views are almost the same. However, our proposed method can also be applied to scenarios where input views are randomly selected and exhibit greater spatial separation. This flexibility stems from our individual construction of MPIs for each input view, and the capacity of each MPI to render novel views consistently. By assuming that the results of rendered novel views by different MPIs remain the same, regions in the novel view overlapping with input views are effectively constrained, producing coherent and reasonable outcomes.}

\paragraph{Training Details.} 
As we discussed above, we construct per-view MPI with 80 planes for each input view, where each MPI is modeled by one independent four or six-layer leakyrelu-MLP with 256 nodes per layer. We set $\gamma$=10 for the spatial coordinate $\mathbf{x}_i^k$ while no position-encoding for the direction vector $\mathbf{d}_i^{(u_t,v_t)}$. The initial learning rate is $5\times10^{-3}$ and then gradually reduce to $1\times10^{-4}$. At the beginning of the training process, we use only the reconstruction loss  
(Eq.~\ref{equ_4.5}) to train the network. After 15 epochs the whole loss function (Eq.~\ref{equ_4.8}) with appearance and depth consistency loss is used, where both $\lambda_{ac}$ and $\lambda_{dc}$ are set to 1. We train our models with the Adam optimizer with randomly sampled 1024 rays in a batch within 50 epochs by a single NVIDIA RTX 3090 GPU. {It takes about 2 hours to train a scene and 10 seconds to render a target view.}

\paragraph{Metrics.}
We evaluate the quality of rendered novel view images with Peak Signal-to-Noise Ratio (PSNR), Structural Similarity Index Measure (SSIM)~\cite{wang2004image}, and Learned Perceptual Image Patch Similarity (LPIPS)~\cite{zhang2018unreasonable}. For easier comparison, we also report the average score by calculating the geometric mean of $10^{-\text{PSNR}/10}$, $\sqrt{1-\text{SSIM}}$ and \text{LPIPS} for the \textit{LLFF} dataset similar to \cite{niemeyer2022regnerf}.

% Table generated by Excel2LaTeX from sheet 'Sheet1'
\begin{table}[t]
	\centering
	\caption{Quantitative comparisons on the \textit{LLFF} dataset. Our proposed method can achieve state-of-the-art performance. {ft indicates the results fine-tuned on each scene individually.}}
    \setlength{\tabcolsep}{1.2mm}
	\begin{tabular}{lcccc}
		\hline
        \hline
		Method & PSNR$\uparrow$  & SSIM$\uparrow$  & LPIPS$\downarrow$ & Average$\downarrow$ \bigstrut\\
		\hline
		NeRF \cite{mildenhall2021nerf}  & 13.34 & 0.373 & 0.451 & 0.255 \bigstrut[t]\\
            {NeX} \cite{wizadwongsa2021nex}  & {17.36} & {0.591} & {0.369} & {0.163} \bigstrut[t]\\
		DietNeRF \cite{jain2021putting} & 14.94 & 0.370  & 0.496 & 0.232 \\
		InfoNeRF \cite{kim2022infonerf} & 14.37 & 0.349 & 0.457 & 0.238 \\
		PixelNeRF{-ft} \cite{yu2021pixelnerf} & 16.17 & 0.438 & 0.438 & 0.217 \\
		SRF{-ft} \cite{chibane2021stereo}   & 17.07 & 0.436 & 0.529 & 0.203 \\
		MVSNeRF{-ft} \cite{chen2021mvsnerf} & 17.88 & 0.584 & 0.327 & 0.157 \\
		GeCoNeRF \cite{kwak2023geconerf} & 18.55 & 0.578 & 0.340  & 0.150 \\
		RegNeRF \cite{niemeyer2022regnerf} & 19.08 & 0.587 & 0.336 & 0.146 \\
        {MixNeRF} \cite{seo2023mixnerf} & {19.27} & {0.629} & {0.336}  & {0.134} \\
        {FlipNeRF} \cite{seo2023flipnerf} & {19.34} & {0.631} & {0.335}  & {0.133} \bigstrut[b]\\
		\hline
		{\textbf{Ours}} & {\textbf{19.45}} & {\textbf{0.659}} & {\textbf{0.310}} & {\textbf{0.127}} \bigstrut\\
		\hline
        \hline
	\end{tabular}%
	\label{tab2:Quantitative comparisons on the LLFF dataset}%
\end{table}%

\subsection{Comparisons with State-of-the-art Methods}
\subsubsection{Results on the \textit{Shiny} Dataset}
We first compare our proposed method with vanilla NeRF \cite{mildenhall2021nerf}, DietNeRF \cite{jain2021putting} and InfoNeRF \cite{kim2022infonerf} on the challenging \textit{Shiny} dataset proposed by \cite{wizadwongsa2021nex} to demonstrate the effectiveness of CMC. We choose 8 real-world scenes from the official shiny and shiny-extended dataset that contain complex view-dependent effects. As shown in Fig.~\ref{fig3: Qualitive comparisons on the shiny dataset.}, NeRF and DietNeRF will overfit to input views, where the estimated geometry (\textit{i.e.}, the depth map) is quite poor. For InfoNeRF, though it can render a more reasonable depth map, it will fail in more complicated scenes such as "Crest". On account that DietNeRF only uses a high-level semantic loss on the image level to realize consistency across different views, it will generate repeated contents on the rendered novel view image. Differently, InfoNeRF takes advantage of ray entropy loss to regularize the seen/unseen views independently, where no cross-views interactions exist. As a result, for some occluded areas in the novel view that don't appear in the input views, it is quite difficult for them to estimate reasonable contents. Instead, our proposed method, \textit{i.e.}, CMC, can render accurate depth maps and novel view images by virtue of a fully utilize of cross-view consistency. As demonstrated in Tab.~\ref{Quantitative comparisons on 8 scenes of the shiny dataset}, CMC can achieve state-of-the-art performance on all the metrics, which reflects the fact that introducing only physical priors would not be strong enough to deal with complex scenes under the few-shot setting, leverage of cross-view consistency will be helpful for obtaining a more accurate geometry estimation. 

\subsubsection{Results on the \textit{LLFF} Dataset}
Similar to many previous works, we also perform experiments on the common \textit{LLFF} dataset against many state-of-the-art methods to demonstrate the superiority of our proposed method. Specifically, we compare our method with pretraining-based methods (\textit{i.e.}, PixelNeRF \cite{yu2021pixelnerf}, SRF \cite{chibane2021stereo}, MVSNeRF \cite{chen2021mvsnerf}), regularization-based methods (\textit{i.e.}, DietNeRF \cite{jain2021putting}, InfoNeRF \cite{kim2022infonerf}, RegNeRF \cite{niemeyer2022regnerf}), pseudo views-based method (\textit{i.e.}, GeCoNeRF \cite{kwak2023geconerf}) and NeRF \cite{mildenhall2021nerf}. 

As verified in Tab.~\ref{tab2:Quantitative comparisons on the LLFF dataset}, our method can still achieve state-of-the-art performance with a big improvement in SSIM. For qualitative comparisons, as shown in Fig.~\ref{fig4: Qualitive comparisons on the llff dataset}, for methods based on pre-trained network such as MVSNeRF, though they can avoid overfitting to input views to some extent, the rendered novel view images would contain unreasonable artifacts due to the domain gap between training dataset and test scenes. Moreover, for input views with quite big disparities, MVSNeRF still falls into overfitting and estimates wrong geometry, as demonstrated by the scene named "Horns". For regularization-based methods such as InfoNeRF, severe artifacts will exist in the generated novel view images. For RegNeRF, the method with the best performance for few-shot novel view synthesis at present, it can overcome the overfitting problem to a large extent by means of depth smoothing regularization and a well-designed sampling annealing strategy. However, RegNeRF still generates some unreasonable geometry and results in discontinuous depth estimation, as demonstrated by the TV and conference table in the scene named "Room". On the contrary, our proposed method can achieve not only photorealistic novel view synthesis but also quite accurate and continuous depth estimation, without any physical priors serving as the regularization term or any hand-crafted complex sampling strategy to avoid overfitting. In other words, our method can realize few-shot novel view synthesis elegantly with lower complexity, which promises many practical applications.

\begin{figure}[t]
	\begin{center} 
		% \rule{0pt}{2in} \rule{.9\linewidth}{0pt}
		\includegraphics[width=1\linewidth]{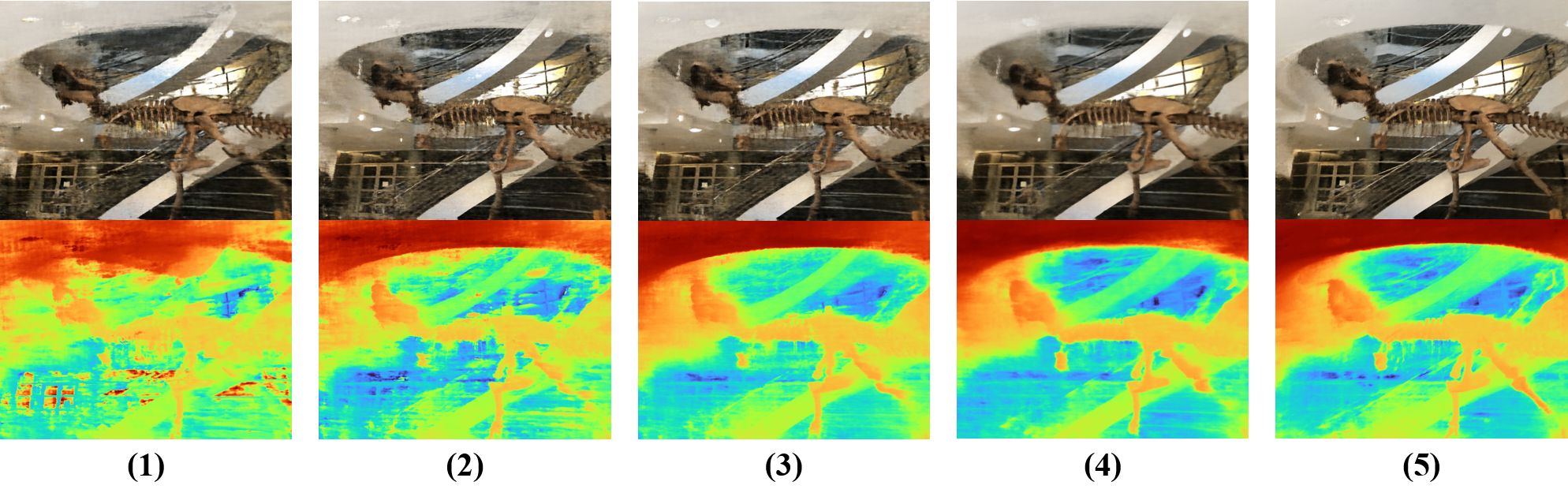}
	\end{center}
	\caption{Qualitative comparisons of different choices of loss functions. (1) Single MPI with $\mathcal{L}_{\text{MSE}}$. (2) Per-view MPI with $\mathcal{L}_{\text{MSE}}$. (3) Per-view MPI with $\mathcal{L}_{\text{MSE}}+\mathcal{L}_{\text{dc}}^{\text{I}}$. (4) Per-view MPI with $\mathcal{L}_{\text{MSE}}+\mathcal{L}_{\text{dc}}^{\text{I}}+\mathcal{L}_{\text{ac}}$. (5) Per-view MPI with $\mathcal{L}_{\text{MSE}}+\mathcal{L}_{\text{dc}}^{\text{I}}+\mathcal{L}_{\text{ac}}+\mathcal{L}_{\text{dc}}$.}
	\label{Qualitative comparisons Ablation studies}
\end{figure}

\subsection{Ablation Studies}
To verify the importance of constructing per-view MPI and the appearance/depth consistency loss, we perform ablation studies on the choices of loss functions. Specifically, we choose loss functions (Eq.~\ref{equ_4.8}) composed of different combinations of the reconstruction loss $\mathcal{L}_{\text{MSE}}$ (Eq.~\ref{equ_4.5}), the depth consistency loss on input views $\mathcal{L}_{\text{dc}}^{\text{I}}$ (Eq.~\ref{equ_4.7}), the appearance consistency loss on novel views $\mathcal{L}_{\text{ac}}$ (Eq.~\ref{equ_4.6}) and the depth consistency loss on novel views $\mathcal{L}_{\text{dc}}$ (Eq.~\ref{equ_4.7}). 

\paragraph{Single MPI.} \label{ablation: single MPI}
As shown in Fig.~\ref{Qualitative comparisons Ablation studies}, for the setting of single MPI (Sing. MPI), \textit{i.e.}, only one random input view is selected as the reference view and thus only one MPI is constructed to render novel views, though some artifacts exist, the neural network can already avoid overfitting to input views. Actually, a single MPI is a variant of NeRF where only the sampling points in rays of different input views are imposed to be distributed on the same planes. However, such a slight change can achieve nearly 4dB PSNR improvement over NeRF as demonstrated in Tab.~\ref{tab2:Quantitative comparisons on the LLFF dataset} and Tab.~\ref{Ablation studies on the choices of different loss functions}. This observation reflects the effectiveness and superiority of our method, where cross-view multiplane consistency can benefit a lot for accurate geometry recovery.

\paragraph{Per-view MPI.} 
To enhance interactions across different input views, we further propose per-view MPI by treating every input view as the reference view and then constructing their corresponding MPIs. As shown in Fig.~\ref{Qualitative comparisons Ablation studies} and Tab.~\ref{Ablation studies on the choices of different loss functions}, with only $\mathcal{L}_{\text{MSE}}$, per-view MPI can witness an increase in rendering quality and generate more accurate geometry estimation, which demonstrates the effectiveness of per-view MPI. Then, when we successively add $\mathcal{L}_{\text{dc}}^{\text{I}}$, $\mathcal{L}_{\text{ac}}$ and $\mathcal{L}_{\text{dc}}$ into the loss function, a continuous growth of performance can be observed, where more photorealistic novel view images and better depth estimation can be achieved. 

% Table generated by Excel2LaTeX from sheet 'Sheet1'
\begin{table}[t]
  \centering
  \caption{Ablation studies on the choices of different loss functions.}
  \setlength{\tabcolsep}{1.2mm}
    \begin{tabular}{c|c|ccccc}
    \hline
    \hline
    Loss  & Sing. MPI & \multicolumn{5}{c}{Per-view MPI} \bigstrut\\
    \hline
    $\mathcal{L}_{\text{MSE}}$ & \ding{51}     & \ding{51}     & \ding{51}     & \ding{51}     & \ding{51}     & \ding{51} \bigstrut[t]\\
    $\mathcal{L}_{\text{dc}}^{\text{I}}$ & \ding{55}      & \ding{55}      & \ding{51}     & \ding{55}      & \ding{51}     & \ding{51} \\
    $\mathcal{L}_{\text{ac}}$ & \ding{55}      & \ding{55}      & \ding{55}      & \ding{51}     & \ding{51}    & \ding{51} \\
    $\mathcal{L}_{\text{dc}}$ & \ding{55}      & \ding{55}      & \ding{55}      & \ding{55}      & \ding{55}      & \ding{51} \bigstrut[b]\\
    \hline
    
    PSNR$\uparrow$  & {17.56} & {18.33} & {18.69} & {19.24} & {19.27} & {\textbf{19.45}} \\
    SSIM$\uparrow$  & {0.597} & {0.618}  & {0.634} & {0.656} & {0.656} & {\textbf{0.659}} \\
    LPIPS$\downarrow$ & {0.359} & {0.345} & {0.334} & {0.336} & {0.321} & {\textbf{0.310}} \\
    Average$\downarrow$ & {0.158} & {0.146} & {0.139} & {0.133} & {0.130} & {\textbf{0.129}} \bigstrut[b]\\
    \hline
    \hline
    \end{tabular}%
  \label{Ablation studies on the choices of different loss functions}%
\end{table}%

\section{Conclusions}
We present a brand-new technique for few-shot novel view synthesis by cross-view multiplane consistency (\textit{i.e.}, CMC). We propose to address the overfitting problem of few-shot view synthesis by forcing the sampling points to be the identical when rendering different views through multiplane images. This is based on the assumption that given sparse input view images, the sampling point in each ray would rarely be used to render other views and thus cause the neural networks to memorize input views rather than learn the underlying geometry. Then, to enhance interactions across different views, we propose to construct per-view MPI by viewing every input view as the reference view followed by leverage of appearance and depth consistency loss. We further provide an explanation for the overfitting problem and give the intuition behind CMC. To verify our assumption and method, we conduct experiments on a large amount of complex real-world scenes, where our proposed CMC can achieve state-of-the-art few-shot novel view synthesis, without any scene priors or complicated hand-crafted sampling strategy.

{\section{Limitations and future works}
\paragraph{Limitations.} 
The main limitation of our proposed method is that CMC doesn't perform well on surrounding scenes that contain big camera rotations, such as the Blender dataset~\cite{mildenhall2021nerf}. This is because CMC is based on Multiplane Images (\textit{i.e.}, MPI), which is specially designed for forward-facing scenes while not suitable to represent surrounding scenes. 

\paragraph{Future works} 
Our future works include extending CMC to surrounding scenes using methods such as multisphere representation, where sampling points are forced to be distributed on the same spheres. Moreover, we will try to use only one MLP instead of per-view MPI to represent the scene, which would decrease the training burden to a large extent.}

{\section{Acknowledgement}
This work was supported in part by NSFC under Grant 62371434, U1908209, 62021001.}

\bibliographystyle{abbrv-doi}

\bibliography{template}

\begin{thebibliography}{10}

\bibitem{ahn2022panerf}
Y.~C. Ahn, S.~Jang, S.~Park, J.-Y. Kim, and N.~Kang.
\newblock Panerf: Pseudo-view augmentation for improved neural radiance fields based on few-shot inputs.
\newblock {\em arXiv preprint arXiv:2211.12758}, 2022.

\bibitem{arpit2017closer}
D.~Arpit, S.~Jastrz{\k{e}}bski, N.~Ballas, D.~Krueger, E.~Bengio, M.~S. Kanwal, T.~Maharaj, A.~Fischer, A.~Courville, Y.~Bengio, et~al.
\newblock A closer look at memorization in deep networks.
\newblock In {\em International conference on machine learning}, pp. 233--242. PMLR, 2017.

\bibitem{blumer1987occam}
A.~Blumer, A.~Ehrenfeucht, D.~Haussler, and M.~K. Warmuth.
\newblock Occam's razor.
\newblock {\em Information processing letters}, 24(6):377--380, 1987.

\bibitem{buehler2001unstructured}
C.~Buehler, M.~Bosse, L.~McMillan, S.~Gortler, and M.~Cohen.
\newblock Unstructured lumigraph rendering.
\newblock In {\em Proceedings of the 28th annual conference on Computer graphics and interactive techniques}, pp. 425--432, 2001.

\bibitem{chaurasia2013depth}
G.~Chaurasia, S.~Duchene, O.~Sorkine-Hornung, and G.~Drettakis.
\newblock Depth synthesis and local warps for plausible image-based navigation.
\newblock {\em ACM Transactions on Graphics (TOG)}, 32(3):1--12, 2013.

\bibitem{chen2021mvsnerf}
A.~Chen, Z.~Xu, F.~Zhao, X.~Zhang, F.~Xiang, J.~Yu, and H.~Su.
\newblock Mvsnerf: Fast generalizable radiance field reconstruction from multi-view stereo.
\newblock In {\em Proceedings of the IEEE/CVF International Conference on Computer Vision}, pp. 14124--14133, 2021.

\bibitem{chen2022geoaug}
D.~Chen, Y.~Liu, L.~Huang, B.~Wang, and P.~Pan.
\newblock Geoaug: Data augmentation for few-shot nerf with geometry constraints.
\newblock In {\em European Conference on Computer Vision}, pp. 322--337. Springer, 2022.

\bibitem{chibane2021stereo}
J.~Chibane, A.~Bansal, V.~Lazova, and G.~Pons-Moll.
\newblock Stereo radiance fields (srf): Learning view synthesis for sparse views of novel scenes.
\newblock In {\em Proceedings of the IEEE/CVF Conference on Computer Vision and Pattern Recognition}, pp. 7911--7920, 2021.

\bibitem{debevec1996modeling}
P.~E. Debevec, C.~J. Taylor, and J.~Malik.
\newblock Modeling and rendering architecture from photographs: A hybrid geometry-and image-based approach.
\newblock In {\em Proceedings of the 23rd annual conference on Computer graphics and interactive techniques}, pp. 11--20, 1996.

\bibitem{deng2022depth}
K.~Deng, A.~Liu, J.-Y. Zhu, and D.~Ramanan.
\newblock Depth-supervised nerf: Fewer views and faster training for free.
\newblock In {\em Proceedings of the IEEE/CVF Conference on Computer Vision and Pattern Recognition}, pp. 12882--12891, 2022.

\bibitem{flynn2019deepview}
J.~Flynn, M.~Broxton, P.~Debevec, M.~DuVall, G.~Fyffe, R.~Overbeck, N.~Snavely, and R.~Tucker.
\newblock Deepview: View synthesis with learned gradient descent.
\newblock In {\em Proceedings of the IEEE/CVF Conference on Computer Vision and Pattern Recognition}, pp. 2367--2376, 2019.

\bibitem{flynn2016deepstereo}
J.~Flynn, I.~Neulander, J.~Philbin, and N.~Snavely.
\newblock Deepstereo: Learning to predict new views from the world's imagery.
\newblock In {\em Proceedings of the IEEE conference on computer vision and pattern recognition}, pp. 5515--5524, 2016.

\bibitem{fridovich2022plenoxels}
S.~Fridovich-Keil, A.~Yu, M.~Tancik, Q.~Chen, B.~Recht, and A.~Kanazawa.
\newblock Plenoxels: Radiance fields without neural networks.
\newblock In {\em Proceedings of the IEEE/CVF Conference on Computer Vision and Pattern Recognition}, pp. 5501--5510, 2022.

\bibitem{han2022single}
Y.~Han, R.~Wang, and J.~Yang.
\newblock Single-view view synthesis in the wild with learned adaptive multiplane images.
\newblock In {\em ACM SIGGRAPH 2022 Conference Proceedings}, pp. 1--8, 2022.

\bibitem{huang2022stylizednerf}
Y.-H. Huang, Y.~He, Y.-J. Yuan, Y.-K. Lai, and L.~Gao.
\newblock Stylizednerf: consistent 3d scene stylization as stylized nerf via 2d-3d mutual learning.
\newblock In {\em Proceedings of the IEEE/CVF Conference on Computer Vision and Pattern Recognition}, pp. 18342--18352, 2022.

\bibitem{jain2021putting}
A.~Jain, M.~Tancik, and P.~Abbeel.
\newblock Putting nerf on a diet: Semantically consistent few-shot view synthesis.
\newblock In {\em Proceedings of the IEEE/CVF International Conference on Computer Vision}, pp. 5885--5894, 2021.

\bibitem{kim2022infonerf}
M.~Kim, S.~Seo, and B.~Han.
\newblock Infonerf: Ray entropy minimization for few-shot neural volume rendering.
\newblock In {\em Proceedings of the IEEE/CVF Conference on Computer Vision and Pattern Recognition}, pp. 12912--12921, 2022.

\bibitem{kwak2023geconerf}
M.~Kwak, J.~Song, and S.~Kim.
\newblock Geconerf: Few-shot neural radiance fields via geometric consistency.
\newblock {\em arXiv preprint arXiv:2301.10941}, 2023.

\bibitem{le2020novel}
H.-A. Le, T.~Mensink, P.~Das, and T.~Gevers.
\newblock Novel view synthesis from single images via point cloud transformation.
\newblock {\em arXiv preprint arXiv:2009.08321}, 2020.

\bibitem{levoy1996light}
M.~Levoy and P.~Hanrahan.
\newblock Light field rendering.
\newblock In {\em Proceedings of the 23rd annual conference on Computer graphics and interactive techniques}, pp. 31--42, 1996.

\bibitem{li2021mine}
J.~Li, Z.~Feng, Q.~She, H.~Ding, C.~Wang, and G.~H. Lee.
\newblock Mine: Towards continuous depth mpi with nerf for novel view synthesis.
\newblock In {\em Proceedings of the IEEE/CVF International Conference on Computer Vision}, pp. 12578--12588, 2021.

\bibitem{li2020crowdsampling}
Z.~Li, W.~Xian, A.~Davis, and N.~Snavely.
\newblock Crowdsampling the plenoptic function.
\newblock In {\em Computer Vision--ECCV 2020: 16th European Conference, Glasgow, UK, August 23--28, 2020, Proceedings, Part I 16}, pp. 178--196. Springer, 2020.

\bibitem{liu2022neural}
Y.~Liu, S.~Peng, L.~Liu, Q.~Wang, P.~Wang, C.~Theobalt, X.~Zhou, and W.~Wang.
\newblock Neural rays for occlusion-aware image-based rendering.
\newblock In {\em Proceedings of the IEEE/CVF Conference on Computer Vision and Pattern Recognition}, pp. 7824--7833, 2022.

\bibitem{mildenhall2019local}
B.~Mildenhall, P.~P. Srinivasan, R.~Ortiz-Cayon, N.~K. Kalantari, R.~Ramamoorthi, R.~Ng, and A.~Kar.
\newblock Local light field fusion: Practical view synthesis with prescriptive sampling guidelines.
\newblock {\em ACM Transactions on Graphics (TOG)}, 38(4):1--14, 2019.

\bibitem{mildenhall2021nerf}
B.~Mildenhall, P.~P. Srinivasan, M.~Tancik, J.~T. Barron, R.~Ramamoorthi, and R.~Ng.
\newblock Nerf: Representing scenes as neural radiance fields for view synthesis.
\newblock {\em Communications of the ACM}, 65(1):99--106, 2021.

\bibitem{mu20223d}
F.~Mu, J.~Wang, Y.~Wu, and Y.~Li.
\newblock 3d photo stylization: Learning to generate stylized novel views from a single image.
\newblock In {\em Proceedings of the IEEE/CVF Conference on Computer Vision and Pattern Recognition}, pp. 16273--16282, 2022.

\bibitem{muller2022instant}
T.~M{\"u}ller, A.~Evans, C.~Schied, and A.~Keller.
\newblock Instant neural graphics primitives with a multiresolution hash encoding.
\newblock {\em ACM Transactions on Graphics (ToG)}, 41(4):1--15, 2022.

\bibitem{niemeyer2022regnerf}
M.~Niemeyer, J.~T. Barron, B.~Mildenhall, M.~S. Sajjadi, A.~Geiger, and N.~Radwan.
\newblock Regnerf: Regularizing neural radiance fields for view synthesis from sparse inputs.
\newblock In {\em Proceedings of the IEEE/CVF Conference on Computer Vision and Pattern Recognition}, pp. 5480--5490, 2022.

\bibitem{park2021nerfies}
K.~Park, U.~Sinha, J.~T. Barron, S.~Bouaziz, D.~B. Goldman, S.~M. Seitz, and R.~Martin-Brualla.
\newblock Nerfies: Deformable neural radiance fields.
\newblock In {\em Proceedings of the IEEE/CVF International Conference on Computer Vision}, pp. 5865--5874, 2021.

\bibitem{pumarola2021d}
A.~Pumarola, E.~Corona, G.~Pons-Moll, and F.~Moreno-Noguer.
\newblock D-nerf: Neural radiance fields for dynamic scenes.
\newblock In {\em Proceedings of the IEEE/CVF Conference on Computer Vision and Pattern Recognition}, pp. 10318--10327, 2021.

\bibitem{rasmussen2000occam}
C.~Rasmussen and Z.~Ghahramani.
\newblock Occam's razor.
\newblock {\em Advances in neural information processing systems}, 13, 2000.

\bibitem{riegler2021stable}
G.~Riegler and V.~Koltun.
\newblock Stable view synthesis.
\newblock In {\em Proceedings of the IEEE/CVF Conference on Computer Vision and Pattern Recognition}, pp. 12216--12225, 2021.

\bibitem{roessle2022dense}
B.~Roessle, J.~T. Barron, B.~Mildenhall, P.~P. Srinivasan, and M.~Nie{\ss}ner.
\newblock Dense depth priors for neural radiance fields from sparse input views.
\newblock In {\em Proceedings of the IEEE/CVF Conference on Computer Vision and Pattern Recognition}, pp. 12892--12901, 2022.

\bibitem{schonberger2016structure}
J.~L. Schonberger and J.-M. Frahm.
\newblock Structure-from-motion revisited.
\newblock In {\em Proceedings of the IEEE conference on computer vision and pattern recognition}, pp. 4104--4113, 2016.

\bibitem{seo2023flipnerf}
S.~Seo, Y.~Chang, and N.~Kwak.
\newblock Flipnerf: Flipped reflection rays for few-shot novel view synthesis.
\newblock In {\em Proceedings of the IEEE/CVF International Conference on Computer Vision}, pp. 22883--22893, 2023.

\bibitem{seo2023mixnerf}
S.~Seo, D.~Han, Y.~Chang, and N.~Kwak.
\newblock Mixnerf: Modeling a ray with mixture density for novel view synthesis from sparse inputs.
\newblock In {\em Proceedings of the IEEE/CVF Conference on Computer Vision and Pattern Recognition}, pp. 20659--20668, 2023.

\bibitem{sinha2009piecewise}
S.~Sinha, D.~Steedly, and R.~Szeliski.
\newblock Piecewise planar stereo for image-based rendering.
\newblock In {\em 2009 International Conference on Computer Vision}, pp. 1881--1888, 2009.

\bibitem{somraj2023simplenerf}
N.~Somraj, A.~Karanayil, and R.~Soundararajan.
\newblock Simplenerf: Regularizing sparse input neural radiance fields with simpler solutions.
\newblock In {\em SIGGRAPH Asia 2023 Conference Papers}, pp. 1--11, 2023.

\bibitem{song2020deep}
Z.~Song, W.~Chen, D.~Campbell, and H.~Li.
\newblock Deep novel view synthesis from colored 3d point clouds.
\newblock In {\em Computer Vision--ECCV 2020: 16th European Conference, Glasgow, UK, August 23--28, 2020, Proceedings, Part XXIV 16}, pp. 1--17. Springer, 2020.

\bibitem{srinivasan2017learning}
P.~P. Srinivasan, T.~Wang, A.~Sreelal, R.~Ramamoorthi, and R.~Ng.
\newblock Learning to synthesize a 4d rgbd light field from a single image.
\newblock In {\em Proceedings of the IEEE International Conference on Computer Vision}, pp. 2243--2251, 2017.

\bibitem{trevithick2021grf}
A.~Trevithick and B.~Yang.
\newblock Grf: Learning a general radiance field for 3d representation and rendering.
\newblock In {\em Proceedings of the IEEE/CVF International Conference on Computer Vision}, pp. 15182--15192, 2021.

\bibitem{truong2021learning}
P.~Truong, M.~Danelljan, L.~Van~Gool, and R.~Timofte.
\newblock Learning accurate dense correspondences and when to trust them.
\newblock In {\em Proceedings of the IEEE/CVF Conference on Computer Vision and Pattern Recognition}, pp. 5714--5724, 2021.

\bibitem{truong2022sparf}
P.~Truong, M.-J. Rakotosaona, F.~Manhardt, and F.~Tombari.
\newblock Sparf: Neural radiance fields from sparse and noisy poses.
\newblock {\em arXiv preprint arXiv:2211.11738}, 2022.

\bibitem{tucker2020single}
R.~Tucker and N.~Snavely.
\newblock Single-view view synthesis with multiplane images.
\newblock In {\em Proceedings of the IEEE/CVF Conference on Computer Vision and Pattern Recognition}, pp. 551--560, 2020.

\bibitem{wang2022nerf}
C.~Wang, R.~Jiang, M.~Chai, M.~He, D.~Chen, and J.~Liao.
\newblock Nerf-art: Text-driven neural radiance fields stylization.
\newblock {\em arXiv preprint arXiv:2212.08070}, 2022.

\bibitem{wang2022generalizable}
D.~Wang, X.~Cui, S.~Salcudean, and Z.~J. Wang.
\newblock Generalizable neural radiance fields for novel view synthesis with transformer.
\newblock {\em arXiv preprint arXiv:2206.05375}, 2022.

\bibitem{wang2021ibrnet}
Q.~Wang, Z.~Wang, K.~Genova, P.~P. Srinivasan, H.~Zhou, J.~T. Barron, R.~Martin-Brualla, N.~Snavely, and T.~Funkhouser.
\newblock Ibrnet: Learning multi-view image-based rendering.
\newblock In {\em Proceedings of the IEEE/CVF Conference on Computer Vision and Pattern Recognition}, pp. 4690--4699, 2021.

\bibitem{wang2004image}
Z.~Wang, A.~C. Bovik, H.~R. Sheikh, and E.~P. Simoncelli.
\newblock Image quality assessment: from error visibility to structural similarity.
\newblock {\em IEEE transactions on image processing}, 13(4):600--612, 2004.

\bibitem{wiles2020synsin}
O.~Wiles, G.~Gkioxari, R.~Szeliski, and J.~Johnson.
\newblock Synsin: End-to-end view synthesis from a single image.
\newblock In {\em Proceedings of the IEEE/CVF Conference on Computer Vision and Pattern Recognition}, pp. 7467--7477, 2020.

\bibitem{wizadwongsa2021nex}
S.~Wizadwongsa, P.~Phongthawee, J.~Yenphraphai, and S.~Suwajanakorn.
\newblock Nex: Real-time view synthesis with neural basis expansion.
\newblock In {\em Proceedings of the IEEE/CVF Conference on Computer Vision and Pattern Recognition}, pp. 8534--8543, 2021.

\bibitem{wood2000surface}
D.~N. Wood, D.~I. Azuma, K.~Aldinger, B.~Curless, T.~Duchamp, D.~H. Salesin, and W.~Stuetzle.
\newblock Surface light fields for 3d photography.
\newblock In {\em Proceedings of the 27th annual conference on Computer graphics and interactive techniques}, pp. 287--296, 2000.

\bibitem{xu2022sinnerf}
D.~Xu, Y.~Jiang, P.~Wang, Z.~Fan, H.~Shi, and Z.~Wang.
\newblock Sinnerf: Training neural radiance fields on complex scenes from a single image.
\newblock In {\em Computer Vision--ECCV 2022: 17th European Conference, Tel Aviv, Israel, October 23--27, 2022, Proceedings, Part XXII}, pp. 736--753. Springer, 2022.

\bibitem{xu2022point}
Q.~Xu, Z.~Xu, J.~Philip, S.~Bi, Z.~Shu, K.~Sunkavalli, and U.~Neumann.
\newblock Point-nerf: Point-based neural radiance fields.
\newblock In {\em Proceedings of the IEEE/CVF Conference on Computer Vision and Pattern Recognition}, pp. 5438--5448, 2022.

\bibitem{yang2023freenerf}
J.~Yang, M.~Pavone, and Y.~Wang.
\newblock Freenerf: Improving few-shot neural rendering with free frequency regularization.
\newblock In {\em Proceedings of the IEEE/CVF Conference on Computer Vision and Pattern Recognition}, pp. 8254--8263, 2023.

\bibitem{you2022learning}
M.~You, M.~Guo, X.~Lyu, H.~Liu, and J.~Hou.
\newblock Learning a unified 3d point cloud for view synthesis.
\newblock {\em arXiv preprint arXiv:2209.05013}, 2022.

\bibitem{yu2021plenoctrees}
A.~Yu, R.~Li, M.~Tancik, H.~Li, R.~Ng, and A.~Kanazawa.
\newblock Plenoctrees for real-time rendering of neural radiance fields.
\newblock In {\em Proceedings of the IEEE/CVF International Conference on Computer Vision}, pp. 5752--5761, 2021.

\bibitem{yu2021pixelnerf}
A.~Yu, V.~Ye, M.~Tancik, and A.~Kanazawa.
\newblock pixelnerf: Neural radiance fields from one or few images.
\newblock In {\em Proceedings of the IEEE/CVF Conference on Computer Vision and Pattern Recognition}, pp. 4578--4587, 2021.

\bibitem{zhang2021understanding}
C.~Zhang, S.~Bengio, M.~Hardt, B.~Recht, and O.~Vinyals.
\newblock Understanding deep learning (still) requires rethinking generalization.
\newblock {\em Communications of the ACM}, 64(3):107--115, 2021.

\bibitem{zhang2020nerf++}
K.~Zhang, G.~Riegler, N.~Snavely, and V.~Koltun.
\newblock Nerf++: Analyzing and improving neural radiance fields.
\newblock {\em arXiv preprint arXiv:2010.07492}, 2020.

\bibitem{zhang2018unreasonable}
R.~Zhang, P.~Isola, A.~A. Efros, E.~Shechtman, and O.~Wang.
\newblock The unreasonable effectiveness of deep features as a perceptual metric.
\newblock In {\em Proceedings of the IEEE conference on computer vision and pattern recognition}, pp. 586--595, 2018.

\bibitem{zhao2022generative}
X.~Zhao, F.~Ma, D.~G{\"u}era, Z.~Ren, A.~G. Schwing, and A.~Colburn.
\newblock Generative multiplane images: Making a 2d gan 3d-aware.
\newblock In {\em Computer Vision--ECCV 2022: 17th European Conference, Tel Aviv, Israel, October 23--27, 2022, Proceedings, Part V}, pp. 18--35. Springer, 2022.

\bibitem{zhou2018stereo}
T.~Zhou, R.~Tucker, J.~Flynn, G.~Fyffe, and N.~Snavely.
\newblock Stereo magnification: learning view synthesis using multiplane images.
\newblock {\em ACM Transactions on Graphics (TOG)}, 37(4):1--12, 2018.

\bibitem{zhou2016view}
T.~Zhou, S.~Tulsiani, W.~Sun, J.~Malik, and A.~A. Efros.
\newblock View synthesis by appearance flow.
\newblock In {\em Computer Vision--ECCV 2016: 14th European Conference, Amsterdam, The Netherlands, October 11--14, 2016, Proceedings, Part IV 14}, pp. 286--301. Springer, 2016.

\end{thebibliography}
\end{document}